\documentclass[final,3p,times,twocolumn]{elsarticle}

\usepackage{amssymb}
\usepackage{amsthm}
\usepackage{amsmath}
\usepackage{multirow}
\usepackage{algorithm}
\usepackage{algorithmic}
\usepackage{color}
\usepackage{verbatim}
\usepackage{diagbox}
\usepackage{bbding}

\usepackage{lineno}

\journal{Springer}

\definecolor{red}{rgb}{1.0, 0.0, 0.0}

\begin{document}

\begin{frontmatter}



\title{TLCE: Transfer-Learning Based Classifier Ensembles for Few-Shot Class-Incremental Learning}


\author[1]{Shuangmei Wang}
\author[1]{Yang Cao}
\affiliation[1]{organization={Jilin University},
    addressline={No. 2699 Qianjin Street}, 
    city={Changchun},
    postcode={130012}, 
    country={China}}
\affiliation[2]{organization={Engineering Research Center of Knowledge-Driven Human-Machine Intelligence},
    city={MOE},
    country={China}}

\author[1,2]{Tieru Wu\corref{cor1}}

\cortext[cor1]{Corresponding authors.}

\begin{abstract}
    Few-shot class-incremental learning (FSCIL) struggles to incrementally recognize novel classes from few examples without catastrophic forgetting of old classes or overfitting to new classes.
    We propose TLCE, which ensembles multiple pre-trained models to improve separation of novel and old classes.
    TLCE minimizes interference between old and new classes by mapping old class images to quasi-orthogonal prototypes using episodic training. It then ensembles diverse pre-trained models to better adapt to novel classes despite data imbalance.
    Extensive experiments on various datasets demonstrate that our transfer learning ensemble approach outperforms state-of-the-art FSCIL methods.

\end{abstract}




\end{frontmatter}


	

\section{Introduction}
    Deep learning has sparked substantial advancements in various computer vision tasks.
    These advancements are mainly due to the emergence of large-scale datasets and powerful GPU computing devices.
    However, deep learning-based methods exhibit limitations in recognizing classes that have not been incorporated into their training.
    In this scenario, there has been significant research conducted on Class-Incremental Learning (CIL), which focuses on dynamically updating the model using only new samples from each additional task, while preserving knowledge about previously learned classes.
    On the other hand, the process of obtaining and annotating a sufficient quantity of data samples presents challenges in both complexity and expense.
    Certain studies are dedicated to investigating CIL in situations where data availability is limited. Specifically, researchers have explored the concept of few-shot class-incremental learning (FSCIL), which aims to continuously learn new classes using only a limited number of target samples.

    As a consequence, two issues arise: the potential for catastrophic forgetting of previously learned classes and the risk of overfitting to new concepts.
    Furthermore, Constrained Few-Shot Class-Incremental Leaning (C-FSCIL) \cite{hersche2022constrained} introduce that this particular learning approach abides by explicit constraints related to memory and computational capacity. 
    These constraints include the necessity to maintain a consistent computational cost when acquiring knowledge about a new class and ensuring that the model's memory usage increases at most linearly as additional classes are introduced.

    To solve the above issues, recent studies \cite{SPRR2021, CEC2021, shi2021F2M} focus on addressing these challenges by emphasizing the acquisition of transferable features through initially utilizing the cross-entropy (CE) loss during training in the base session, while also subsequently freezing the backbone to facilitate adaptation to new classes.
    C-FSCIL \cite{hersche2022constrained} employs meta-learning to map input images to quasi-orthogonal prototypes in a way that minimizes interference between the prototypes of different classes.
    Although C-FSCIL has demonstrated superior performance, we find a prediction bias arising from class imbalance and data imbalance.
    We also observe that the process of assigning hyperdimensional quasi-orthogonal vectors to each class demands a substantial number of samples and iterations. 
    This undoubtedly presents a challenge when it comes to allocating prototypes to novel classes that possess only a limited amount of samples.
    
    In this paper, we propose TLCE, a transfer-learning based few-shot class-incremental learning method that ensembles various classifiers memorized different knowledge.
    One main inspiration is pretraining a deep network on the base dataset and transferring knowledge to the novel classes \cite{closer, rethinking} has been shown as the strong baseline for the few-shot classification. 
    On the other hand, little interference between the new classes and the old classes is key.
    Hence, we leverage the advantages offered by the aforementioned classifiers through ensemble learning.
    Firstly, we employ meta-learning to train a robust hyperdimensional network (RHD) according to C-FSCIL. 
    This allows us to effectively map input images to quasi-orthogonal prototypes for base classes.
    Secondly, we integrate cosine similarity and cross-entropy loss to train a transferable knowledge network (TKN).
    Finally, we compute the prototype, i.e., the average of features, for each class. 
    The classification of a test sample is simply determined by finding its nearest  prototype measured by the weighted integration combines the different relationships.

    Comparing to C-FSCIL, our TLCE adopts the similar idea of assigning quasi-orthogonal prototypes for base classes to reduce minimal interference.
    The key difference is the attempt to perform well on all classes equally, regardless of the training sequence employed through classifier ensembles.
    We conduct extensive comparisons with state-the-art few-shot class-incremental classificaiton methods on miniImageNet \cite{2015ImageNet} and CIFAR100 \cite{cifar2009learning} and the results demonstrate the superiority of our TLCE. Ablation studies on different ensembles, i.e., different weights between the robust hyperdimensional network and transferable knowledge network also show the necessity of ensembling two classifiers for better results.
    
    In summary, our contributions are as follows:

    \begin{enumerate}
		\item We propose TLCE, transfer-learning based classifier ensembles to improve the novel class set separation and maintain the base class set separation.
		\item  Without additional training and expensive computation, the proposed method can efficiently explore the comprehensive relation between prototypes and test features and  improve the novel class set separation and maintain the base class set separation.
		\item We conduct extensive experiments on various datasets and the results show our efficient method can outperform SOTA few-shot class-incremental classification methods.
	\end{enumerate}
\section{Related Work}
    
    \textbf{Few-Shot Learning.}
    
    FSL seeks to develop neural models for new categories using only a small number of labeled samples. 
    Meta-learning \cite{hospedales2021meta} is extensively utilized to accomplish few-shot classification.
    The core idea is to use the episodic training paradigm to learn generalizable classifiers or feature extractors for the data of the base classes in an optimization-based framework \cite{MAML,Jamal_2019_CVPR,LEO}, as well as learn a distance function to measure the similarity among feature embeddings through metric-learning \cite{koch2015siamese,matching,prototypical, relation}.
    On the other hand, pretraining classifiers or image encoders on the base dataset and then adapting them the novel classes via transfer learning \cite{closer,rethinking} has been shown as the strong baseline for the few-shot classification.
    Based on the meta-learned feature extractor or the pretrained deep image model, we can perform nearest-neighbor (NN) based classification which has been proven as a simple and effective approach for FSL. 
    Specially, the prediction is determined by measuring the similarity or distance between the test feature and the prototypes of the novel labeled features.
    Due to the limited number of samples, the prototypes computed from the few-shot novel class data may cannot represent the underlying data distribution.
    Several methods \cite{DC, guo2022learning, xu2022alleviating, wang2023p3dc} have been proposed to perform data calibration to obtain better samples or prototypes of the novel class recently.
    Inspired by those representative few-shot methods, we attempt to leverage different training paradigms to acquire diverse models to calculate target prototypes for the few-shot class-incremental learning tasks.

    \textbf{Class Incremental Learning.}
    CIL aims to build a universal classifier among all seen classes from a stream of labeled training sets. 
    Current CIL algorithms can be roughly divided into three categories. 
    The first category utilizes former data for rehearsal, which enables the model to review former instances and overcome forgetting \cite{iscen2020memory, zhu2021prototype, petit2023fetril}.
    The second category estimates the importance of each parameter and keeps the important ones static \cite{kirkpatrick2017overcoming, chaudhry2018riemannian, lee2020continual}.
    Other methods designs algorithms to maintain the model's knowledge and discriminability.
    For example, knowledge distillation-based methods build the mapping between old and new models \cite{rebuffi2017icarl, eeil2018, gao2022r-dfcil}. On the other hand, several methods aim to find bias and rectify them like the oracle model \cite{yu2020semantic, liu2021adaptive, xu2022alleviating}. FSCIL can be seen a particular case of the CIL. Therefore, we can learn from some of the above methods. 
    
    \textbf{Few-Shot Class-Incremental Learning.}
    FSCIL introduces few-shot scenarios where only a few labeled samples are available into the task of class-incremental learning. To achieve FSCIL, many works attempt to solve the problem of catastrophic forgetting and seriously overfitting from different perspective. 
    TOPIC \cite{Tao2020} employs a neural gas network to preserve the topology of the feature manifold from a cognitive-inspired perspective.
    SKD \cite{cheraghian2021semantic} and ERDIL \cite{dong2021few} use knowledge distillation to to balance the preserving of old-knowledge and adaptation of new-knowledge.
    Feature-space based methods focus on obtaining compact clustered features and maintaining generalization for future incremental classes \cite{ zhou2022forward, peng2022few, song2023learning}.
    From the perspective of parameter space, WaRP \cite{kim2023warping} combines the advantages of F2M \cite{shi2021F2M} to find flat minimums of the loss function and FSLL \cite{2021FSLL} for parameter fine-tuning.
    They push most of the previous knowledge compactly into only a few important parameters so that they can fine-tune more parameters during incremental sessions.
    From the perspective of hybrid approaches, some works combine episodic training \cite{CEC2021, chi2022metafscil, hersche2022constrained}, ensemble learning \cite{ji2023memorizing, xu2023flexible}, and so on.
    C-FSCIL \cite{hersche2022constrained} maps input images to quasi-orthogonal prototypes such that the prototypes of different classes encounter small interference through episodic training.
    However, achieving quasi-orthogonality among all prototypes for the classes poses difficulties when dealing with novel classes that have only a limited number of labeled samples.
    MCNet \cite{ji2023memorizing} trains multiple embedding networks using diverse network architectures to to enhance the diversity of models and enable them to memorize different knowledge effectively.
    Similar to the above method, our method is based on ensemble learning, while we train two shared architecture networks using different loss function and training methods.
    \cite{xu2023flexible} enhances the expression ability of extracted features through multistage pre-training and uses meta-learning process to extract meta-feature as complementary features. 
    Please note that a novel generalization model is one with no overlapping among novel class sets and no interference with base classes.
    In contrast to these methods, we ensemble a robust hyperdimensional (HD) network for base classes and a trasnferable knowledge network for novel classes from a whole new perspective.

\section{Method}
\begin{figure*}[ht]
		\centering
		\includegraphics[width=1.0\linewidth]{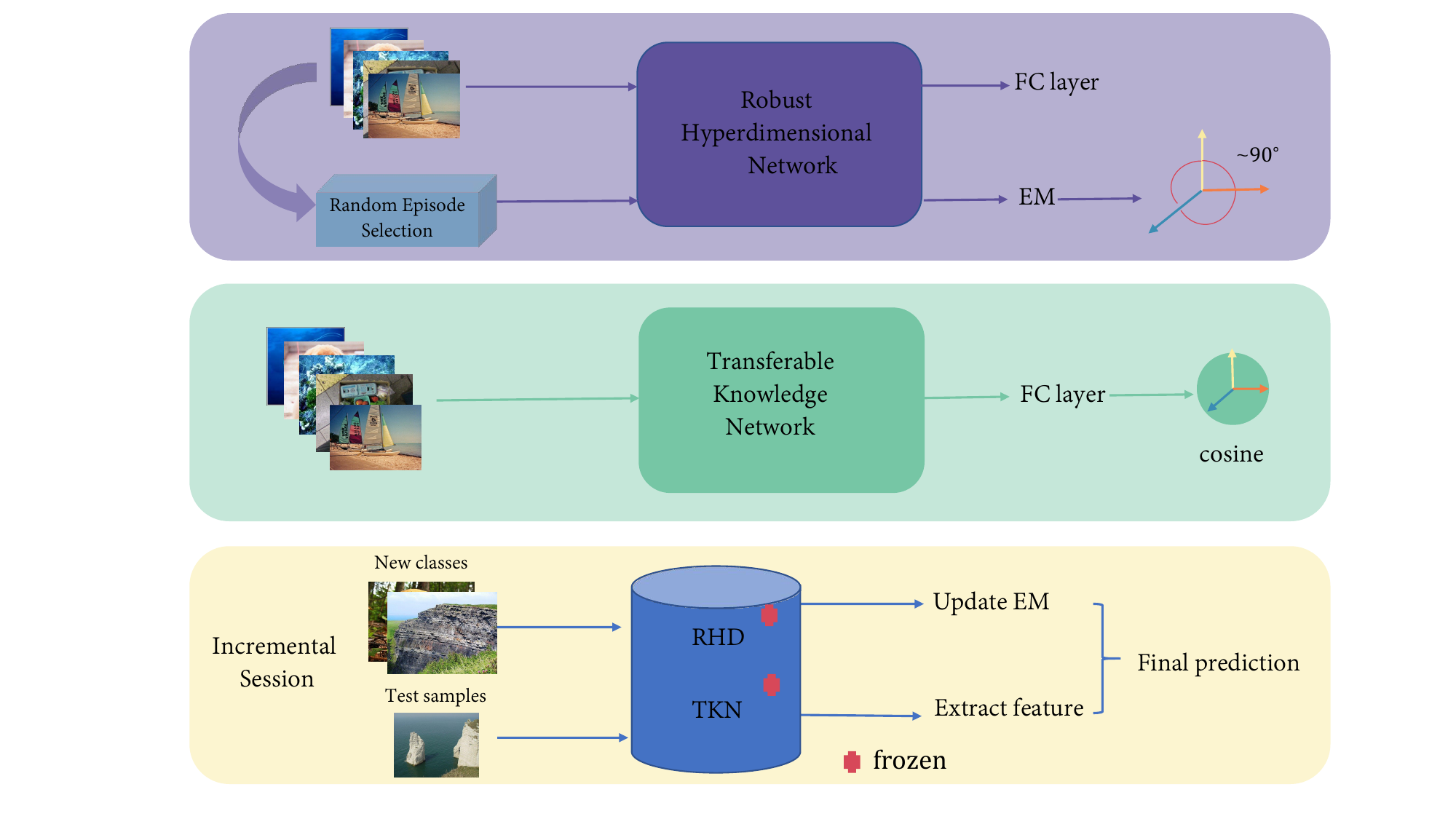}
		\caption{An illustration of the proposed method pipeline. $F$ is the backbone network and $G$ is a projection layer. The RHD and TKN have a shared architecture.
        We obtained different network parameters by using various training methods and loss functions. 
        In the incremental session, we freeze the RHD and TKN parameters.
		}
		\label{framework}
	\end{figure*}

    In this section, we propose the FSCIL method using model ensemble. 
    An ideal FSCIL learning model should ensure that the newly added categories do not interfere with the old ones and maintain a distinct separation between them.
    The motivations mentioned above prompt us to solve the aforementioned problems by combining a robust hyperdimensional memory-augmented neural network and a transferable knowledge model through ensemble.
    Firstly, we draw inspiration from \cite{karunaratne2021robust, hersche2022constrained} and employ episodic training to map the base datasets to quasi-orthogonal prototypes, thereby minimizing interference of base classes during incremental sessions.
    Secondly, we pretrain a model from scratch in a standard supervised way to gain transferable knowledge space.
    Finally, we have integrated explicit memory (EM) into the previously mentioned embedding networks. 
    This has been done in a manner that allows the EM to store the embeddings of labeled data samples as class prototypes within its memory. 
    During the testing process, we utilize the nearest prototype classification method based on similarity thereby meeting the classification requirements for all seen classes.
    Note that we only need to compute the new class prototypes using the aforementioned models and update the EM because training only takes place within the base session.
    Figure \ref{framework} demonstrates the framework of our method.
    In the following, we provide technical details of the proposed method for few-shot class-incremental classification.

    \subsection{Problem Statement}
    FSCIL learn continuously from a sequential stream of tasks.
    Suppose we have a constant stream of labeled training data denoting $D^1, D^2, \dots, D^T$, where $D^t= \{(x_i, y_i)\}_{i=1}^{|D^t|}$.
    In the t-th task, the set of labels is denoted as $C^t$, where where $\forall i\neq j, C^i \cap C^j=  \emptyset $.
    The total number of classes in this task is represented by  $\lvert C^t \rvert$. 
    $D^1$ with ample data is called the base session, while  $D^t$ (where $t > 1$) pertains to the limited training set involving new classes (called incremental session).
    We follow the conventional few-shot class-incremental learning setting, i.e., build a series of N-way K-shot datasets $D^t = { \{(x_i, y_i)\}_{i=1}^{N\times K}}$, where N is the number of novel classes and K is the number of port samples in each novel session. 
    For each session $t$, the model only have access to the dataset $D^t$.
    After training with $D^t$, the model needs to recognize all encountered classes in $\cup_{s\leq t}C^s$.

    \subsubsection{Robust Hyperdimensinal Network (RHD)}
    Due to the "curse" of dimensionality, a randomly selected vector has a high probability of being quasi-orthogonal to other random vectors. 
    As a result, when representing a novel class, the process not only contributes incrementally to previous learning but also causes minimal interference.
    Hence, we follow C-FSCIL \cite{hersche2022constrained} to build a RHD network during the base session.
    
    Our method is comprised of three primary components: a backbone network , an extra projection, and a fully connected layer.
    The backbone network maps the samples from the input domain $\mathcal{X}$ to a feature space.
    In order to construct an embedding network that utilizes a high-dimensional distributed representation, the backbone network is joined with a projection layer.
    Then we have 
    \begin{equation}
        \label{eq:robust}
        \mu_1 = F_{\theta_1} (x ), \quad \mu_2 = G_{\theta_2}(\mu_1),
    \end{equation}
    where $\mu_1 \in R^{d_f}$ is the intermediate feature of input $x$, $d_f $ is the dimension of the feature space, $\mu_2 \in R^{d}$ is the output feature of the intermediate feature $\mu_1$, and $\theta_1 ,\theta_2$ are the learnable parameters of the backbone network and the projection layer, respectively.
    
    Firstly, we jointly train both $F_{\theta_1}$ and $G_{\theta_2}$ from scratch in the standard supervised classification using the base session data to derive powerful embeddings for the downstream base learner.
    The empirical risk to minimize can be formulated as:
    \begin{equation}
	\min_{\theta_1, \theta_2} \quad
    L_{ce} ((W^T\mu_2), y),
    \end{equation}
    where $L_{ce} ( \cdot , \cdot )$ is cross-entropy loss (CE) and $W^T$ is the learnable parameters of the fully connected layer.

    Lastly, we build on top of the meta-learning setup to allocate nearly quasiorthogonal vectors to various image classes. These vectors are then positioned far away from each other in the hyperdimensional space.
    We replace the fully connected layer with the EM and build a series of $\lvert D^1 \rvert $-way K-shot tasks where $\lvert D^1 \rvert $ is the number of base classes and K is the number of support samples in each task.
    In every task, the projection layer produces a support vector for every training input. 
    To represent each class, we calculate the average of all support vectors that belong to a specific class, thereby generating a single prototype vector for that class.
    Within the EM, prototypes are saved for each class. Specifically, the prototype for a given class $i$ is determined in the following manner:
    \begin{equation}
         \label{eq:p}
        p_i^R =  \frac{1}{ \lvert \mathbf{S}_i \rvert}\sum_{x \in \mathbf{S}_i} G_{\theta_2}(F_{\theta_1}(x)), 
    \end{equation}
    where $\mathbf{S}_i$ is the set of all samples from class $i$ and $ \lvert S_c \rvert$ is the number of samples.
    Given a query sample $q$ and prototypes, we compute the cosine similarity for class $i$ as follows:
    \begin{equation}
		\label{eq:similarity}
		S_i^R = cos(tanh(G_{\theta_2}(F_{\theta_1}(q)), tanh(p_i^R)),
	\end{equation}
    where $\tanh(\cdot)$ is the hyperbolic tangent function and $cos(\cdot, \cdot)$ is the cosine similarity.
    In hyperdimensional memory-augmented neural networks \cite{karunaratne2021robust}, the hyperbolic tangent has demonstrated its usefulness as a non-linear function that regulates the activated prototypes' norms and embedding outputs. 
    Additionally, cosine similarity tackles the norm and bias problems commonly encountered in FSCIL by emphasizing the angle between activated prototypes and embedding outputs while disregarding their norms \cite{lesort2021continual}. 
    Given the cosine similarity score $S_i^R$ for every class $i$, we utilize a soft absolute sharpening function to enhance this attention vector, resulting in quasi-orthogonal vectors \cite{karunaratne2021robust}.
    \textbf{Softabs attention} 
    The softabs attention function is defined as 
    \begin{align}
        h(S_i^R) = \frac{\epsilon(S_i^R)}{\sum_{j=1}^{|D^1|}  \epsilon(S_j^R)}, 
    \end{align}
    where $\epsilon(\cdot)$ is the sharpening function: 
    \begin{align}
        \epsilon(c) = \frac{1}{1+e^{-(\beta (c-0.5))}} + \frac{1}{1+e^{-(\beta (-c-0.5))}}. 
    \end{align}
    The sharpening function includes a stiffness parameter $\beta$, which is set to 10 as in~\cite{karunaratne2021robust}.

    \subsubsection{Transferable Knowledge Network (TKN)}
    It is difficult to ensure quasi-orthogonality among all prototypes for each class due to the presence of novel classes that only have a small number of labeled samples.
    Inspired by transfer learning based few-shot methods, we explore various transferable models.
    The most straightforward approach involves utilizing a model that has been pre-trained from the scratch using standard supervised classification techniques. We employ this model as a baseline for our analysis.
    
    In SimpleShot \cite{simpleshot}, it demonstrates that using nearest neighbor classification, where features are simply normalized by L2 norm and measured by Euclidean distance, can obtain competitive results in few-shot classification tasks.
    The squared Euclidean distance after L2 normalization is equivalent to cosine similarity.
    Utilizing cosine similarity as a distance metric for quantifying data similarity has two implications: 1) during training, it focuses on the angles between normalized features rather than the absolute distances within the latent feature space, and 2) the normalized weight parameters of the fully connected layer can be interpreted as the centroids or centers of each category \cite{peng2022few}. 
    So we combine cosine similarity with cross-entropy loss to train a more transferable network.
    To simplify calculations of cosine similarity in the final fully connected layer, we set the bias to zero.
    Then the data prediction procedure can be written as:
	\begin{equation}
	\mu_2 = G_{\theta_2}(\mu_1) = G_{\theta_2}(F_{\theta_1}(x)),
	\label{eq: feature_extractor}
	\end{equation}
	\begin{equation}
	w_i = W_i^T  \mu_2 = \|W_i\| \|\mu_2\| \cos(\theta_i) = \cos(\theta_i), \nonumber
	\end{equation}
	\begin{equation}
	\|W_i\| = \|\mu_2\| = 1.
	\end{equation}
	\label{eq: cosine_similarity}
    The quantity $w_i$ is the calculated cosine similarity between the feature $\mu_2$ and the weight parameter $W_i$ for class i.
    The loss function is given by:
    \begin{equation}
    \begin{aligned}
	L 
    &= -\frac{1}{T}\sum_{j=1}^{T}\log(\frac{e^{y_{j}}}{\sum^{\lvert C^1 \rvert}_{i=1}e^{w_i}})\\
	& = -\frac{1}{T}\sum_{j=1}^{T}\log(\frac{e^{\|W_{j}\| \|\mu_2\| \cos(\theta_{j})}}{\sum_{i=1}^{\lvert C^1 \rvert}e^{\|W_i\| \|\mu_2\| \cos(\theta_i)}})\\
	& = -\frac{1}{T}\sum_{j=1}^{T}\log(\frac{e^{\cos(\theta_{j})}}{\sum_{i=1}^{\lvert C^1 \rvert}e^{\cos(\theta_i)}}),\\
	\end{aligned}
	\label{eq: cross_entropy}
	\end{equation}
    where T is the number of training images and the quantity $y_j$ describes the cosine similarity towards its ground truth class for image j.

    \subsection{Incremental Test}
    By employing the incremental-frozen framework, we can reduce the storage requirements by only preserving the prototypes of all the encountered classes and updating the exemplar memory (EM) when introducing new classes. 
    This way, we can effectively manage the limitations imposed by memory and computational capacities.
    Firstly, we utilize the robust hyperdimensional network and transferable knowledge network to calculate the prototypes $P^R$ and  $P^T$. 
    Once we acquire the prototypes for the novel classes, we can promptly update the EM.
    It is important to note that the EM does not update the prototypes of the old classes, as RHD and TKN remain fixed in the subsequent session.
    Then, we save all the prototypes for the classes that have been appeared so far within the EM.
    Finally, we can derive the ultimate classification outcome by evaluating the similarity measure between the test sample and each prototype.
    Suppose we have a test sample $q$.
    According to Eq. \ref{eq:similarity}, we can compute separate similarity $S^R$ and $S^T$ for each classifier RHD and TKN individually.
    Then, we can combine classifiers through weighted integration by considering both scores to obtain the final score $S$ as:
   
    \begin{equation}
         \begin{aligned}
		\label{eq:ensemble}
		&S = \lambda *S^R +(1-\lambda) S^T,
	    \end{aligned}
        \end{equation}
    where $\lambda \in [0, 1]$ is the hyperparameter.
    This approach allows us to leverage the strengths of multiple classifiers and intelligently merge their outputs, leading to a more accurate final result.

    \begin{table*}[t]
    \caption{Quantitative comparison on the test set of miniImageNet in the 5-way 5-shot FSCIL setting. "Average Acc." is the average performance of all sessions.  "Final Improv." calculates the improvement of our method in the last session.  $\dagger$ The results of \cite{hersche2022constrained} are obtained using its released code.}
    \centering
    \setlength{\tabcolsep}{4.0pt}

    \begin{tabular}{l|ccccccccccc}
     \multirow{2}{*}{\textbf{Method}}
     & \multicolumn{9}{c}{\textbf{Session ID}} 
     & \textbf{Average}
     & \textbf{Final}\\
     \cline{2-10} 
     &  \textbf{1} & \textbf{2} &\textbf{3}
     &\textbf{4} & \textbf{5}& \textbf{6}&\textbf{7} & \textbf{8}& \textbf{9} & \textbf{Acc.} & \textbf{Improv.}
    \\
    \hline
    
    iCaRL \cite{rebuffi2017icarl} & 61.31 & 46.32 & 42.94 & 37.63 & 30.49 & 24.00 & 20.89 & 18.80 & 17.21 & 33.29 & \textbf{+35.79} \\

     EEIL \cite{eeil2018} & 61.31 & 46.58 & 44.00 & 37.29 & 33.14 & 27.12 & 24.10 & 21.57 & 19.58 & 34.97 & \textbf{+33.42} \\

     NCM \cite{NCM2019} & 61.31 & 47.80 & 39.30 & 31.90 & 25.70 & 21.40 & 18.70 & 17.20 & 14.17 & 30.83 & \textbf{+38.83} \\
    
    TOPIC \cite{Tao2020} & 61.31 & 50.09 & 45.17 & 41.16 & 37.48 & 35.52 & 32.19 & 29.46 & 24.42 & 39.64 & \textbf{+28.58} \\
    
    SKD \cite{cheraghian2021semantic} & 61.31 & 58.00 & 53.00 & 50.00 & 48.00 & 45.00 & 42.00 & 40.00 & 39.00 &48.48 &\textbf{+14.00} \\

    SPPR \cite{SPRR2021} & 61.45 & 63.80 & 59.53 & 55.53 & 52.50 & 49.60 & 46.69 & 43.79 & 41.92 &52.76 &\textbf{+11.08} \\

    F2M\cite{shi2021F2M} & 72.05 & 67.47 & 63.16 & 59.70 & 56.71 & 53.77 & 51.11 & 49.21 & 47.84 &57.89 &\textbf{+5.16}\\
    
    CEC\cite{CEC2021} & 72.00 & 66.83 & 62.97 & 59.43 & 56.70 & 53.73 & 51.19 & 49.24 & 47.63 & 57.75 &\textbf{+5.37} \\
    
    C-FSCIL$\dagger$\cite{hersche2022constrained} & 76.38 & 70.77 & 66.17 & 62.67& 59.17 &  56.2 & 53.27&  51.09 & 48.93&60.52 &\textbf{+4.07} \\

    \hline
    
    \textbf{Ours(baseline)} & \textbf{76.88} & \textbf{71.83} & \textbf{67.46} &  \textbf{ 64.39} & \textbf{ 61.29} & \textbf{ 58.47} & \textbf{55.79} &  \textbf{ 53.81}  &\textbf{52.30} & \textbf{62.47} &\textbf{+0.70}\\
    
    \textbf{Ours} & \textbf{77.05} & \textbf{71.72} & \textbf{67.51} &  \textbf{64.40} & \textbf{61.55} & \textbf{58.76} & \textbf{56.27} &  \textbf{54.35}  &\textbf{53.00} & \textbf{62.73} &\\
    \hline

    \end{tabular}
    \vspace{-0.15in}
    \label{table:mini}
    \end{table*}

    \begin{table*}[t]
    \caption{Quantitative comparison on the test set of CIFAR100 in the 5-way 5-shot FSCIL setting. "Average Acc." is the average performance of all sessions.  "Final Improv." calculates the improvement of our method in the last session. $\dagger$ The results of \cite{hersche2022constrained} are obtained using its released code.}
    \centering
    \setlength{\tabcolsep}{4.0pt}
   
    \begin{tabular}{l|ccccccccccc}
     \multirow{2}{*}{\textbf{Method}}
     & \multicolumn{9}{c}{\textbf{Session ID}} 
     & \textbf{Average}
     & \textbf{Final}\\
     \cline{2-10} 
     &  \textbf{1} & \textbf{2} &\textbf{3}
     &\textbf{4} & \textbf{5}& \textbf{6}&\textbf{7} & \textbf{8}& \textbf{9} & \textbf{Acc.} & \textbf{Improv.}
    \\
    \hline
    iCaRL \cite{rebuffi2017icarl} & 64.10 & 53.28 & 41.69 & 34.13 & 27.93 & 25.06 & 20.41 & 15.48 & 13.73 & 32.87 & \textbf{+38.98}\\

     EEIL \cite{eeil2018} & 64.10 & 53.11 & 43.71 & 35.15 & 28.96 & 24.98 & 21.01 & 17.26 & 15.85 & 33.79 & \textbf{+36.86} \\

     NCM \cite{NCM2019} & 64.10 & 53.05 & 43.96 & 36.97 & 31.61 & 26.73 & 21.23 & 16.78 & 13.54 & 34.22 & \textbf{+39.17} \\
    \hline
    
    TOPIC \cite{Tao2020} & 64.10 & 55.88 & 47.07 & 45.16 & 40.11 & 36.38 & 33.96 & 31.55 & 29.37 & 42.62 & \textbf{+23.34} \\
   
    SKD \cite{cheraghian2021semantic} & 64.10 & 57.00 & 50.01 & 46.00 & 44.00 & 42.00 & 39.00 & 37.00 & 35.00 & 46.01 & \textbf{+17.71}\\

    SPPR \cite{SPRR2021} & 64.10 & 65.86 & 61.36 & 57.45 & 53.69& 50.75 & 48.58 & 45.66 & 43.25 &54.52 &\textbf{+9.46} \\

    F2M\cite{shi2021F2M} & 71.45 & 68.10 & 64.43 & 60.80 & 57.76 & 55.26 & 53.53 & 51.57 & 49.35 &59.14 &\textbf{+3.36} \\
    
    CEC\cite{CEC2021} & 73.07 & 68.88 & 65.26 & 61.19 & 58.09 & 55.57 & 53.22 & 51.34 & 49.14 & 59.53  &\textbf{+3.57}\\

    C-FSCIL$\dagger$\cite{hersche2022constrained} & 77.22 & 71.92 & 67.16 & 63.01 & 59.21 & 56.13 & 53.44 & 51.05 & 48.94 & 60.90 & \textbf{+3.77} \\

    \hline
   
   \textbf{Ours(baseline)} &  \textbf{77.33} &  \textbf{72.49} & \textbf{68.30} &  \textbf{64.25} &  \textbf{60.89} & \textbf{58.18} &  \textbf{55.68} & \textbf{53.65}  & \textbf{51.63} & \textbf{62.49} &\textbf{+1.08} \\

    \textbf{Ours} & \textbf{77.82} & \textbf{73.00} & \textbf{69.11} &  \textbf{65.17} & \textbf{62.29} & \textbf{59.48} & \textbf{56.96} &  \textbf{54.97}  &\textbf{52.71} & \textbf{63.50}& \\
    \hline
    
    \end{tabular}
    \vspace{-0.15in}
    \label{table:cifar}
    \end{table*}
\section{Experiments}
 
     In this section, we conduct quantitive comparisons between our TLCE and state-of-the-art few-shot class-incremental learning methods on two representative datasets. We also perfrom ablation studies on evaluating design choices and different hyperparameters for our methods.
    \subsection{Datasets}
    We evaluate our proposed method on two datasets for benchmarking few-shot class-incremental learning: miniImageNet \cite{2015ImageNet} and CIFAR100 \cite{cifar2009learning}. 

    In the miniImageNet \cite{2015ImageNet} dataset, there are 100 classes, with each class having 500 training images and 100 testing images. 
    As for CIFAR100 \cite{cifar2009learning}, it is a challenging dataset with 60,000 images of size 32 × 32, divided into 100 classes. 
    Each class has 500 training images and 100 testing images.
    Following the split used in \cite{Tao2020}, we select 60 base classes and 40 novel classes from CIFAR100 and miniImageNet. 
    These 40 novel classes are further divided into eight incremental sessions. 
    In each session, we learn using a 5-way 5-shot approach, which means training on 5 classes with 5 images per class.
	
    \subsection{Implementation Details}
    For miniImageNet and CIFAR100, we use ResNet-12 following C-FSCIL \cite{hersche2022constrained}.
    We train the TKN with the SGD optimizer, where the learning rate is 0.01, the batch size is set as 128 and epoch is 120.
    As for the RHD network, it is pretrained by the C-FSCIL \cite{hersche2022constrained} work.
    For each image in the dataset, we represent it as a 512-dimensional feature extracor.
    The hyperparameter $\lambda$ is set to 0.8 for both the miniImageNet and the CIFAR100 dataset.

    \subsection{Comparison and Evaluation}
       \begin{figure*}[h]
    		\centering
		\includegraphics[width=0.95\linewidth]{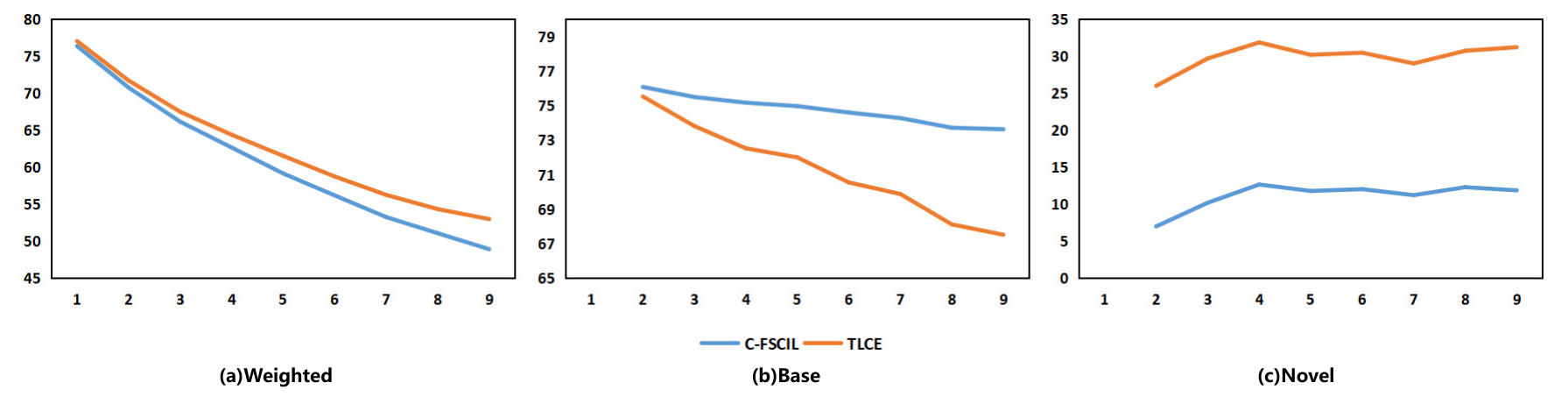}
		\caption{The weighted, base, and novel performances on miniImageNet.
		}
		\label{compare}
	\end{figure*}
 \begin{table*}[t]
		\caption{The ablation study on value selection of hyperparameter $\lambda$. Accuracy (\%) on the test set of miniImageNet and CIFAR100 in the last session are measured.}
		\begin{center}
  \setlength{\tabcolsep}{4.0pt}
			\scalebox{1.0}{
				\begin{tabular}{l|cccccccccccc}
					\hline
					\diagbox{Datasets}{$\lambda$}                            & 0.0 & 0.1 & 0.2 & 0.3 & 0.4 & 0.5 & 0.6 & 0.7& 0.8& 0.9 & 1.0\\
					\hline
					
					miniImageNet  &48.93 &49.21 &49.57 &50.14 &50.83 &51.53 &52.29 &52.78 &53.00 &51.17 &44.87 \\
					
					CIFAR100  &48.94 &49.02 &49.74 &50.30 &50.80 &51.46 &52.42 &52.61 &52.71 &50.80 &44.96 \\   
					\hline

				\end{tabular}
			}
		\end{center}
		\label{hyperparameter}
	\end{table*}

  \begin{table*}[t]
		\caption{The effect of various components of ensemble classifiers.
        Accuracy (\%) on the test set of miniImageNet are measured.}
		\begin{center}
			\scalebox{0.95}{
				\begin{tabular}{c|c| c| c ccccccccc}
					\hline
				Cross Entropy & Cosine & RHD        & 1 & 2 & 3 & 4 & 5 & 6 & 7 & 8 &9\\
					\hline

                \Checkmark &  & &71.10 &64.14 &59.14 &55.92 &53.44 &50.86 &47.97 &46.23 &44.85  \\
                \hline
					
                \Checkmark &\Checkmark &
                &70.55 &63.35 &58.63 &55.53 &53.09 &50.39 &47.72 &46.04 &44.87\\
                \hline
                & & \Checkmark &  76.38 & 70.77 & 66.17 & 62.67& 59.17 &  56.2 & 53.27&  51.09 & 48.93 \\
					\hline
     
			  \Checkmark  & &\Checkmark 
            & \textbf{76.88} & \textbf{71.83} & \textbf{67.46} &  \textbf{ 64.39} & \textbf{ 61.29} & \textbf{ 58.47} & \textbf{55.79} &  \textbf{ 53.81}  &\textbf{52.30}\\
                \hline
                \Checkmark & \Checkmark & \Checkmark & \textbf{77.05} & \textbf{71.72} & \textbf{67.51} &  \textbf{64.40} & \textbf{61.55} & \textbf{58.76} & \textbf{56.27} &  \textbf{54.35}  &\textbf{53.00} \\   
					\hline

				\end{tabular}
			}
		\end{center}
		\label{ensemble}
	\end{table*}

    In order to evaluate the effectiveness of our TLCE, we first conduct quantitative comparisons with several representative and state-of-art few-shot classs-incremental learning methods.
    However, it is important to note that an improvement does not necessarily imply an improvement in both base and novel performances individually.
    Then, we conduct further analysis of model
    performance from both perspectives base and novel to delve deeper into the performance improvement.
    Furthermore, our method offers the advantage of requiring no additional training and consuming minimal storage space.

    \textbf{Quantitative comparisons.}
    As there are numerous efforts have been paid to the few-shot class-incremental learning, we mainly compare our TLCE with representative and SOTA works.
    The compared methods include CIL methods \cite{rebuffi2017icarl, eeil2018, NCM2019} and FSCIL methods \cite{Tao2020, cheraghian2021semantic, SPRR2021, shi2021F2M, CEC2021, hersche2022constrained}.
    For C-FSCIL \cite{hersche2022constrained}, we only compare with their basic version and do not take their model requiring additional training during incremental sessions into consideration .

    For our method, we report we provide our best results with the value of $\lambda$ set to 0.8.
    Table \ref{table:mini} and \ref{table:cifar} show the quantitative comparison results on two datasets. It can be seen that our best results outperform the other methods.
    In particular, we consider the different transferable knowledge models. 
    For the baseline, we train the model in the standard supervised classification. 
    For TLCE, we integrate cosine metric with cross entropy to train the model .
    It can be seen that the latter one can significantly enhance the performance of the ensemble classifiers.
    
    From Table \ref{table:mini} and \ref{table:cifar}, it can be deserved that C-FSCIL performs more effectively in the first five incremental sessions, while the effectiveness is slight in the last four incremental sessions. 
    We make further analysis from the perspective of the accuracy on base and novel classes, respectively.
    According to the data shown in Figure \ref{compare}, we can observe a slight decrease in the base performance.
    This indicates that C-FSCIL could resist the knowledge forgetting.
    However, the novel performance on the following incremental sessions is poor.
    In contrast, an ideal FSCIL classifier will have equally high performance on both novel and base classes.
    For our method TLCE, it is evident that while there is a decrease in the base classes, there is a significant improvement in the novel and weighted performance.
    In the ablation study, we perform more experiments and analysis of different $\lambda$ values to reveal which degree of RHD and TKN is more suitable for the dataset.

    \subsection{Ablation Study}

    In this section, we perform ablation studies to verify the design choices of our method and the effectiveness of different modules. 
    First, we conduct experiments on different hyperparameter $\lambda$ to see how the RHD and TKN can affect the final results.
    Then, we perform the study on the effectiveness of different ensemble classifiers.

    \textbf{Effect on different hyperparameter $\lambda$.}
    Different $\lambda$ values correspond to different degrees of RHD and TKN applied to the input data.
    From the results in Table \ref{hyperparameter}, it can be found when the TKN does not work ($\lambda = 0.0$), the result is lower. 
    But with the ensemble of TKN, the result shows a convex curve with different $\lambda$. 
    That indicates the importance of the TKN.

    \textbf{Effect on different ensemble classifiers.}
    We conduct experiments on miniImageNet to verify the effectiveness of the ensemble classifiers.
    Specifically, we train the TKN in the standard supervised classifier as the baseline.
    The results in Table \ref{ensemble} show that the ensemble classifier can lead to better performance.
    Furthermore, we discover that integrating cosine metric with cross entropy can lead to further enhancement of model performance.
    Hence, we adopt the latter approach for TKN training in our classification.

\section{Conclusion}
    In this paper, we propose a simple yet effective framework, named TLCE, for few-shot class-incremental learning.
    Without any retraining and expensive computation during incremental sessions, our transfer-learning based ensemble classifiers method can efficiently to further alleviate the issues of catastrophic forgetting and overfitting.
    Extensive experiments show that our method can outperform SOTA methods.
    Investigating a more transferable network is worthy to explore in the future.
    Also, exploring a more general way to combine the classifiers is an interesting future work.

	
	
	
	\bibliographystyle{elsarticle-num-names}

	\bibliography{cas-refs}
	
		
		
		
\end{document}